\documentclass[preprint,12pt,numbers,sort,compress]{elsarticle}

\usepackage{amssymb}
\usepackage{amsmath}

\usepackage{natbib}
\usepackage{multirow}
\usepackage[normalem]{ulem}            
\useunder{\uline}{\ul}{} 
\usepackage{booktabs}
\usepackage{rotating}
\usepackage{comment}
\usepackage{color}
\usepackage{graphicx}
\usepackage{url}
\usepackage{algorithm} 
\usepackage{algorithmic} 
\usepackage[justification=centering]{caption}
\usepackage{subcaption}
\usepackage{array}
\newcolumntype{L}[1]{>{\raggedright\let\newline\\\arraybackslash\hspace{0pt}}m{#1}}
\newcolumntype{C}[1]{>{\centering\let\newline\\\arraybackslash\hspace{0pt}}m{#1}}
\newcolumntype{R}[1]{>{\raggedleft\let\newline\\\arraybackslash\hspace{0pt}}m{#1}}

\usepackage[capitalize,noabbrev]{cleveref}

\usepackage{xcolor}

\journal{Knowledge-Based Systems}

\begin{document}

\begin{frontmatter}

\title{DAMEL: Dual-Axis Multi-Expert Learning for Class-Imbalanced Learning}

\author[inst1]{Hyuck Lee\fnref{equal}}
\ead{dlgur0921@kaist.ac.kr}

\author[inst2]{Taemin Park\fnref{equal}}
\ead{ptm1001@kaist.ac.kr}

\author[inst2]{Heeyoung Kim}
\ead{heeyoungkim@kaist.ac.kr}

\fntext[equal]{Equal contribution}

\affiliation[inst1]{organization={AI Research, Krafton},  
            addressline={Teheran-ro 231, Centerfield EAST, Gangnam-gu},  
            city={Seoul},  
            postcode={06142},  
            country={Republic of Korea}}

\affiliation[inst2]{organization={Department of Industrial and Systems Engineering, Korea Advanced Institute of Science and Technology (KAIST)},  
            addressline={291 Daehak-ro, Yuseong-gu},  
            city={Daejeon},  
            postcode={34141},  
            country={Republic of Korea}}

\begin{abstract}
Various algorithms have been proposed to address the challenges \textcolor{black}{of} class-imbalanced learning \textcolor{black}{in} real-world \textcolor{black}{datasets} with long-tailed distributions. \textcolor{black}{Although} these algorithms reduce prediction bias through rebalancing techniques, they often \textcolor{black}{do so at the cost of} increased prediction variance. Several multi-expert learning algorithms \textcolor{black}{have been developed} to \textcolor{black}{mitigate} this variance\textcolor{black}{; however, they typically} involve complex procedures. We propose a new multi-expert learning algorithm, \textcolor{black}{termed} dual-axis multi-expert learning (DAMEL), \textcolor{black}{that} reduces both \textcolor{black}{prediction} bias and variance by \textcolor{black}{leveraging} multiple experts along \textcolor{black}{the} representation and time axes. Along the representation axis, DAMEL concatenates the representations \textcolor{black}{generated by} multiple experts and \textcolor{black}{simultaneously} trains an auxiliary balanced classifier \textcolor{black}{using} the concatenated \textcolor{black}{representations}. Along the time axis, DAMEL aggregates network weights across training epochs \textcolor{black}{and uses these} aggregated weights during testing. Experimental results demonstrate that DAMEL \textcolor{black}{effectively} reduces both \textcolor{black}{prediction} bias and variance, highlighting its effectiveness \textcolor{black}{for} class-imbalanced learning.
\end{abstract}

\begin{keyword}
Long-tailed learning, Multi-expert learning, Exponential moving average of weights.
\end{keyword}

\end{frontmatter}

\newpage
\section{Introduction}\label{intro}
Deep neural networks have demonstrated remarkable performance across a wide range of tasks owing to their ability to capture complex patterns and \textcolor{black}{efficiently} perform large-scale computations \cite{kim2021locally,kim2023contextual,yoon2024uncertainty}. However, many real-world datasets exhibit class imbalance, which poses a significant challenge \textcolor{black}{when} training classification models \textcolor{black}{and} often \textcolor{black}{results in prediction} bias toward majority classes \cite{choy2016looking,das2022supervised,park2022prediction,lee2023semi,cho2023prediction,yoon2026multimodal}. To address this issue, \textcolor{black}{various} class-imbalanced learning (CIL) methods have been proposed.
Existing CIL techniques include rebalancing class \textcolor{black}{distributions through} resampling \cite{barandela2003restricted,chawla2002smote,japkowicz2000class,he2009learning,lee2023resampling,park2024rebalancing} and reweighting \cite{NIPS2013_9aa42b31,NIPS2017_147ebe63,huang2016learning,zhao2022siamese,lee2025learnable}, decoupled learning \cite{kang2019decoupling, NEURIPS2020_2ba61cc3}, and data augmentation \cite{kim2020m2m, zhang2017mixup, chou2020remix}. These techniques have successfully improved classification accuracy.

However, \textcolor{black}{recent studies have} reported that CIL techniques tend to increase \textcolor{black}{prediction} variance \textcolor{black}{while} reducing the squared bias of predictions \cite{wang2021longtailed}. To \textcolor{black}{mitigate this increased} variance, ensemble strategies \textcolor{black}{that} involve training multiple experts have been proposed \cite{zhou2020bbn,xiang2020learning,cai2021ace,wang2021longtailed}. By reducing \textcolor{black}{prediction} variance, multi-expert learning algorithms \textcolor{black}{have} achieved higher classification accuracy \textcolor{black}{than} other CIL techniques. 
However, \textcolor{black}{these methods often} require multi-stage \textcolor{black}{training involving} knowledge distillation or \textcolor{black}{partitioning} data into multiple groups for \textcolor{black}{individual experts}, making \textcolor{black}{their} implementation \textcolor{black}{relatively complex}.

In this paper, we propose a new multi-expert learning algorithm for CIL, \textcolor{black}{termed} dual-axis multi-expert learning (DAMEL), \textcolor{black}{which leverages} multiple experts along \textcolor{black}{both} the representation and time axes in a conceptually simple manner. Specifically, DAMEL \textcolor{black}{reduces} both the squared bias and variance of predictions compared \textcolor{black}{with} existing multi-expert learning algorithms \textcolor{black}{through} the following \textcolor{black}{two key modifications}: \textcolor{black}{(}1) whereas \textcolor{black}{existing} multi-expert learning algorithms aggregate the predictions of multiple experts, DAMEL aggregates \textcolor{black}{their} representations and trains an auxiliary classifier \textcolor{black}{using the aggregated representations} with a class-balanced loss; \textcolor{black}{and (}2) DAMEL aggregates the neural network \textcolor{black}{weights} at the end of each \textcolor{black}{training} epoch, \textcolor{black}{effectively forming} an ensemble along the time axis. DAMEL can be trained in \textcolor{black}{a single} stage without \textcolor{black}{requiring} complex \textcolor{black}{procedures} such as knowledge distillation or complicated \textcolor{black}{partitioning} of \textcolor{black}{the} training data.

\begin{table*}[t]
\caption{Summary of recent multi-expert learning algorithms\textcolor{black}{.}}
\begin{center}
\resizebox{5.25in}{!}{%
\begin{tabular}{ccccc}
\hline
\toprule
Algorithm & Data split & One-stage/Multi-stage & Ensemble of experts & Time axis ensemble \\
\midrule
BBN \cite{zhou2020bbn}   & O & One-stage learning& Classifier & X \\
\midrule
LFME \cite{xiang2020learning}& O & Multi-stage learning& Classifier & X \\
\midrule
RIDE \cite{wang2021longtailed} & X & Multi-stage learning & Classifier & X\\
\midrule
ACE \cite{cai2021ace}& O & One-stage learning & Classifier & X\\
\midrule
TLC \cite{li2022trustworthy}& X & One-stage learning & Classifier & X\\
\midrule
ResLT \cite{cui2022reslt}& X & One-stage learning & Classifier & X\\
\midrule
DAMEL (Ours)    & X & One-stage learning & Representations & O        \\
\bottomrule
\end{tabular}}
\end{center}

\label{difference}
\end{table*}

Specifically, the multiple experts \textcolor{black}{in} DAMEL share a deep CNN \textcolor{black}{backbone} up to the \textcolor{black}{final} residual block to reduce computational cost, similar to previous multi-expert learning \textcolor{black}{algorithms} \cite{wang2021longtailed, cai2021ace}. \textcolor{black}{Each expert has an independent final residual block and classification layer}, \textcolor{black}{both} of which \textcolor{black}{are} trained using the cross-entropy loss. It is \textcolor{black}{well established} that \textcolor{black}{individual experts can learn} different representations even \textcolor{black}{when} trained with the same loss \textcolor{black}{function} because different \textcolor{black}{network weight} initializations \textcolor{black}{can} lead to \textcolor{black}{distinct} modes \textcolor{black}{during} optimization \cite{fort2019deep,lakshminarayanan2017simple,nam2021diversity}. To \textcolor{black}{leverage} the \textcolor{black}{diverse} representations \textcolor{black}{learned by} multiple experts \textcolor{black}{across these} modes, we concatenate the \textcolor{black}{representations from the final} layers of \textcolor{black}{all} experts and train an auxiliary classifier \textcolor{black}{using} a class-balanced loss \textcolor{black}{on} the concatenated representations. Training the auxiliary classifier with \textcolor{black}{these} aggregated representations reduces both the squared bias and the variance of the predictions because the classifier can \textcolor{black}{exploit diverse information learned by} all experts. \textcolor{black}{In contrast} to \textcolor{black}{the} recent trend \textcolor{black}{of decoupling representation} learning \textcolor{black}{from} classifier \textcolor{black}{learning}, we train the auxiliary classifier and multiple experts end-to-end, \textcolor{black}{allowing} the classifier \textcolor{black}{to} be \textcolor{black}{jointly optimized} with the \textcolor{black}{learned} representations \cite{lee2021abc,wang2021contrastive}. 
In Section~\ref{methodology}, we explain why aggregating representations is \textcolor{black}{more effective} than aggregating the predictions of multiple experts.

In addition to aggregating representations, we aggregate the network weights at the end of each epoch to reduce variability in the \textcolor{black}{learned} weights and, consequently, the variance of the predictions. \textcolor{black}{Specifically, we use an} exponential moving average (EMA) \textcolor{black}{to} aggregate the weights at the end of each epoch, \textcolor{black}{assigning greater importance to network weights obtained in later training epochs. We perform weight aggregation} at \textcolor{black}{the epoch level rather than at every} iteration because \textcolor{black}{ensembles} of predictions \textcolor{black}{are known to become} more effective \textcolor{black}{when} the \textcolor{black}{correlations} among \textcolor{black}{individual} predictions \textcolor{black}{are} lower \cite{mehta2019high}. The EMA weights are automatically updated \textcolor{black}{using} the weights of the trained network \textcolor{black}{and are used} to predict the class labels of new \textcolor{black}{samples during} the test phase. \textcolor{black}{Aggregating} the weights using EMA \textcolor{black}{reduces prediction} variance \textcolor{black}{by introducing an ensemble} effect along an additional axis. \textcolor{black}{Moreover, using} EMA weights \textcolor{black}{can reduce} the squared bias of the predictions because the aggregated weights stably approach local optima \cite{izmailov2018averaging} in the test\textcolor{black}{-}error \textcolor{black}{landscape under} a constant learning rate. The main \textcolor{black}{differences} between DAMEL and other multi-expert learning algorithms are summarized in \textcolor{black}{Table}~\ref{difference}.

Our experimental results on CIFAR-10-LT, CIFAR-100-LT \cite{cui2019class}, ImageNet-LT \cite{liu2019large}, and iNaturalist2018 \cite{van2018inaturalist} under various \textcolor{black}{settings} demonstrate that DAMEL achieves \textcolor{black}{the highest} \textcolor{black}{classification} accuracy in CIL \textcolor{black}{tasks}. Through extensive experimental analysis, we verify that \textcolor{black}{aggregating the} representations of multiple experts\textcolor{black}{, together with aggregating network} weights\textcolor{black}{,} reduces both the squared bias and variance of predictions compared \textcolor{black}{with} existing multi-expert learning algorithms. We also \textcolor{black}{evaluate} the contribution of each component of DAMEL through ablation \textcolor{black}{studies}. The main contributions of this paper are summarized as follows:
\begin{itemize}
\item We introduce a representation-axis ensemble that aggregates expert representations \textcolor{black}{through} concatenation rather than averaging representations or aggregating predictions, \textcolor{black}{enabling} the auxiliary classifier to exploit complementary features learned by different experts.
\item We \textcolor{black}{demonstrate} that representation concatenation in DAMEL is not merely an increase in feature dimensionality but \textcolor{black}{rather} a mechanism for preserving expert-specific representation diversity and jointly \textcolor{black}{leveraging} discriminative cues captured by multiple experts.
\item We introduce a time-axis ensemble based on epoch-level EMA weight aggregation, which differs from conventional iteration-level EMA and stochastic weight averaging (SWA) by treating temporally aggregated weights as an additional ensemble axis for reducing prediction variance.
\item We design an auxiliary balanced classifier \textcolor{black}{that is} trained on concatenated expert representations \textcolor{black}{using} a class-balanced loss, \textcolor{black}{thereby reducing} prediction bias toward majority classes while leveraging diverse expert features \textcolor{black}{within a single}-stage learning \textcolor{black}{framework}.
\item \textcolor{black}{Through} extensive experiments and ablation studies\textcolor{black}{, we demonstrate} that the proposed representation-axis and time-axis ensembles jointly reduce both squared bias and prediction variance, \textcolor{black}{resulting in} improved performance on long-tailed \textcolor{black}{classification} benchmarks.
\end{itemize}

\section{Related Work}

\textbf{Class-Imbalanced Learning:} Traditional approaches to \textcolor{black}{addressing} class imbalance include resampling and reweighting techniques. 
 Resampling \textcolor{black}{methods} \cite{barandela2003restricted,chawla2002smote,japkowicz2000class,he2009learning} balance the class distribution by adjusting the \textcolor{black}{number} of \textcolor{black}{training samples} in each class, \textcolor{black}{whereas} reweighting \textcolor{black}{methods} \cite{NIPS2013_9aa42b31,NIPS2017_147ebe63,huang2016learning,zhao2022siamese} assign \textcolor{black}{greater} weights to minority class \textcolor{black}{samples} during training. \textcolor{black}{Although} these \textcolor{black}{approaches} are \textcolor{black}{relatively} simple to implement, they \textcolor{black}{may} suffer from overfitting, information loss, and training \textcolor{black}{instability} \cite{NEURIPS2019_621461af,an2021why}. To \textcolor{black}{mitigate} these \textcolor{black}{limitations}, Cui  \cite{cui2019class} proposed a reweighted loss based on \textcolor{black}{the} effective number of \textcolor{black}{samples}, \textcolor{black}{while} Jamal  \cite{jamal2020rethinking} and Ren  \cite{ren2018learning} \textcolor{black}{introduced} meta-learning based reweighting \textcolor{black}{strategies}. Cao  \cite{NEURIPS2019_621461af} and Ren  \cite{NEURIPS2020_2ba61cc3} \textcolor{black}{developed new loss functions designed to minimize an upper bound on the} generalization error. Kim  \cite{kim2020m2m} and Yin  \cite{yin2018feature} transferred knowledge \textcolor{black}{learned} from majority class data to minority classes. Kang  \cite{kang2019decoupling} decoupled \textcolor{black}{representation} learning \textcolor{black}{from classifier} training\textcolor{black}{, while} Menon  \cite{menon2020long} proposed a logit-adjusted loss \textcolor{black}{that} is Fisher consistent for balanced error. Wei  \cite{wei2022open} and Jada  \cite{zada2022pure} \textcolor{black}{augmented} minority class data using out-of-distribution data and pure noise images, respectively.  \textcolor{black}{More recently}, contrastive learning \cite{wang2021contrastive,kang2020exploring,cui2021parametric,li2022targeted}, class-balanced distillation \cite{iscen2021class,li2021self}, and multi-expert learning \cite{zhou2020bbn,xiang2020learning,cai2021ace,wang2021longtailed,li2022trustworthy} have \textcolor{black}{attracted considerable} attention. \textcolor{black}{In this study, we} focus on multi-expert learning. 

\textbf{Multi-Expert Learning:} Various multi-expert learning algorithms have been proposed to reduce \textcolor{black}{prediction variance}. BBN \cite{zhou2020bbn} \textcolor{black}{jointly} trains an over-balanced classifier and an imbalanced classifier and aggregates \textcolor{black}{their} predictions. LFME \cite{xiang2020learning} assigns an expert to each non-overlapping class group and \textcolor{black}{combines} the predictions of multiple experts through knowledge distillation. RIDE \cite{wang2021longtailed} \textcolor{black}{employs} an \textcolor{black}{expert-specific} classification loss \textcolor{black}{together with a} distribution-aware diversity loss to \textcolor{black}{promote diversity among} multiple experts. ACE \cite{cai2021ace} \textcolor{black}{assigns} diverse but overlapping class \textcolor{black}{subsets to individual experts}, \textcolor{black}{enabling them to provide complementary} and \textcolor{black}{mutually supporting predictions}. TLC \cite{li2022trustworthy} estimates \textcolor{black}{the} uncertainty of each expert and assigns \textcolor{black}{greater} importance to \textcolor{black}{more reliable} experts \textcolor{black}{during testing}. ResLT \cite{cui2022reslt} \textcolor{black}{uses} a residual fusion mechanism \textcolor{black}{consisting of} a main branch for all classes and two residual branches \textcolor{black}{designed} to \textcolor{black}{improve performance on} medium and tail \textcolor{black}{class samples}. \textcolor{black}{Unlike} these \textcolor{black}{methods, which} aggregate the predictions of multiple experts, \textcolor{black}{our approach aggregates their} representations. In addition, we ensemble the neural network \textcolor{black}{weights} along the time axis using EMA weights \textcolor{black}{during testing}.

\textbf{Exponential Moving Average of Weights:} Recently, EMA weights have been \textcolor{black}{adopted} in various deep learning algorithms. For example,  Tarvainen and Valpola  \cite{tarvainen2017mean} used EMA weights for consistency regularization \textcolor{black}{in} semi-supervised learning\textcolor{black}{, while} He  \cite{he2020momentum} \textcolor{black}{employed them} to \textcolor{black}{construct} a consistent memory module for contrastive learning. Tan  \cite{tan2019mnasnet}\textcolor{black}{,} Howard  \cite{howard2019searching}\textcolor{black}{, and} Tan and Le \cite{tan2019efficientnet} used EMA weights to stabilize the optimization process\textcolor{black}{, whereas} Fan  \cite{fan2022cossl} \textcolor{black}{applied them} to \textcolor{black}{construct} a momentum encoder \textcolor{black}{for} imbalanced semi-supervised learning. Unlike \textcolor{black}{these} studies, we interpret EMA \textcolor{black}{weight aggregation} as \textcolor{black}{an ensemble} along the time axis and analyze \textcolor{black}{its effects on} squared bias and \textcolor{black}{prediction} variance \textcolor{black}{in detail}. \textcolor{black}{Moreover}, whereas \textcolor{black}{previous} studies \textcolor{black}{typically} update EMA weights \textcolor{black}{at every} iteration, we update them \textcolor{black}{at the end of} each epoch to increase the diversity \textcolor{black}{among} the \textcolor{black}{network} weights \textcolor{black}{being} aggregated.

\textbf{Stochastic Weight Averaging (SWA):} Izmailov  \cite{izmailov2018averaging} proposed SWA, which averages \textcolor{black}{network weights} along the time axis to stabilize the optimization process and \textcolor{black}{move} the weights closer to local optima \textcolor{black}{on} the test loss surface. SWA is similar to EMA \textcolor{black}{weight aggregation} in that \textcolor{black}{both methods average network} weights \textcolor{black}{over} time\textcolor{black}{. However}, \textcolor{black}{unlike EMA,} SWA \textcolor{black}{assigns equal importance to all aggregated weights and therefore} cannot \textcolor{black}{place greater emphasis on} weights \textcolor{black}{obtained during later training} epochs. \textcolor{black}{As demonstrated in} Section \ref{ablation}, EMA \textcolor{black}{weight aggregation} is more effective than SWA.

\section{Methodology}
\label{methodology}
\subsection{Problem Setting}
\label{setting}
Suppose that we have a training set $\mathcal{D}=\left\lbrace\left(x_n,y_n\right):n\in\left(1,\ldots,N\right)\right\rbrace$, where $x_n\in\mathbb{R}^d$ \textcolor{black}{denotes} the $n$th data point and $y_n\in\left\lbrace1,\ldots,L\right\rbrace$ \textcolor{black}{denotes its} corresponding label. \textcolor{black}{Let} $N_l$ denote the number of data points belonging to class $l$, \textcolor{black}{such that} $\sum_{l=1}^{L}N_l=N$. \textcolor{black}{Without} loss of generality, we assume that the $L$ classes are sorted in descending order \textcolor{black}{of cardinality}, i.e., $N_1\geq N_2\geq\cdots\geq N_L$. We \textcolor{black}{define} the class\textcolor{black}{-}imbalance \textcolor{black}{ratio} as $\gamma=N_1/N_L$\textcolor{black}{, where} $\gamma\gg1$ \textcolor{black}{in} class-imbalanced \textcolor{black}{settings}.

From $\mathcal{D}$, we \textcolor{black}{construct} minibatches $\mathcal{MB}=\left\lbrace\left(x_b,y_b\right):b\in\left(1,\ldots,B\right)\right\rbrace\subset\mathcal{D}$ \textcolor{black}{at} each \textcolor{black}{training} iteration, where $B$ \textcolor{black}{denotes} the minibatch size. Using these minibatches, \textcolor{black}{our goal is} to learn a classification \textcolor{black}{function} $f:\mathbb{R}^d\rightarrow\left\lbrace1,\ldots,L\right\rbrace$ that performs effectively on a class-balanced test set $\mathcal{D}_{\mathrm{test}}=\left\lbrace\left(x_m,y_m\right):m\in\left(1,\ldots,M\right)\right\rbrace$.

It is well known that there is a trade-off between the squared bias and the variance of the predictions\textcolor{black}{. However}, \textcolor{black}{ensemble strategies} can reduce \textcolor{black}{prediction} variance without \textcolor{black}{necessarily} increasing squared bias. Specifically, if we assume that the class-imbalanced training set $\mathcal{D}$ is randomly drawn from the underlying true distribution, \textcolor{black}{then} $\mathcal{D}$ can \textcolor{black}{be treated} as a random variable. \textcolor{black}{The expected} squared prediction error\textcolor{black}{,} ${\mathbb{E}_{\mathcal{D}}}[\sum_{m=1}^{M}(y_m-f_{\mathcal{D}}(x_m))^2]$\textcolor{black}{,} can \textcolor{black}{then} be decomposed into the squared bias, $\sum_{m=1}^{M}(y_m-{\mathbb{E}_{\mathcal{D}}}[f_{\mathcal{D}}(x_m)])^2$\textcolor{black}{, the} variance\textcolor{black}{,} ${\mathbb{E}_{\mathcal{D}}}[\sum_{m=1}^{M}(f_{\mathcal{D}}(x_m)-{\mathbb{E}_{\mathcal{D}}}[f_{\mathcal{D}}(x_m)])^2]$, and \textcolor{black}{an} irreducible error \textcolor{black}{term} \cite{kong2023flexible}, where $f_{\mathcal{D}}$ \textcolor{black}{denotes} a classification algorithm trained \textcolor{black}{on a} randomly drawn training set $\mathcal{D}$. \textcolor{black}{An} ensemble of multiple experts can \textcolor{black}{achieve greater} variance \textcolor{black}{reduction} as the correlation \textcolor{black}{among their} predictions decreases and the number of experts increases. To effectively ensemble multiple experts without \textcolor{black}{substantially increasing} computational complexity, we aggregate the representations \textcolor{black}{produced by} multiple experts and aggregate the \textcolor{black}{network} weights at the end of each epoch. As described in \textcolor{black}{Sections}~\ref{trainingauxiliary} and~\ref{ema}, our ensemble strategy also reduces the squared bias of the predictions. 
\subsection{Learning \textcolor{black}{Diverse} Representations}
\label{learningvarious}
\begin{figure*}[tb]
	\centering
	\includegraphics[width=1\textwidth, height=2.1in]{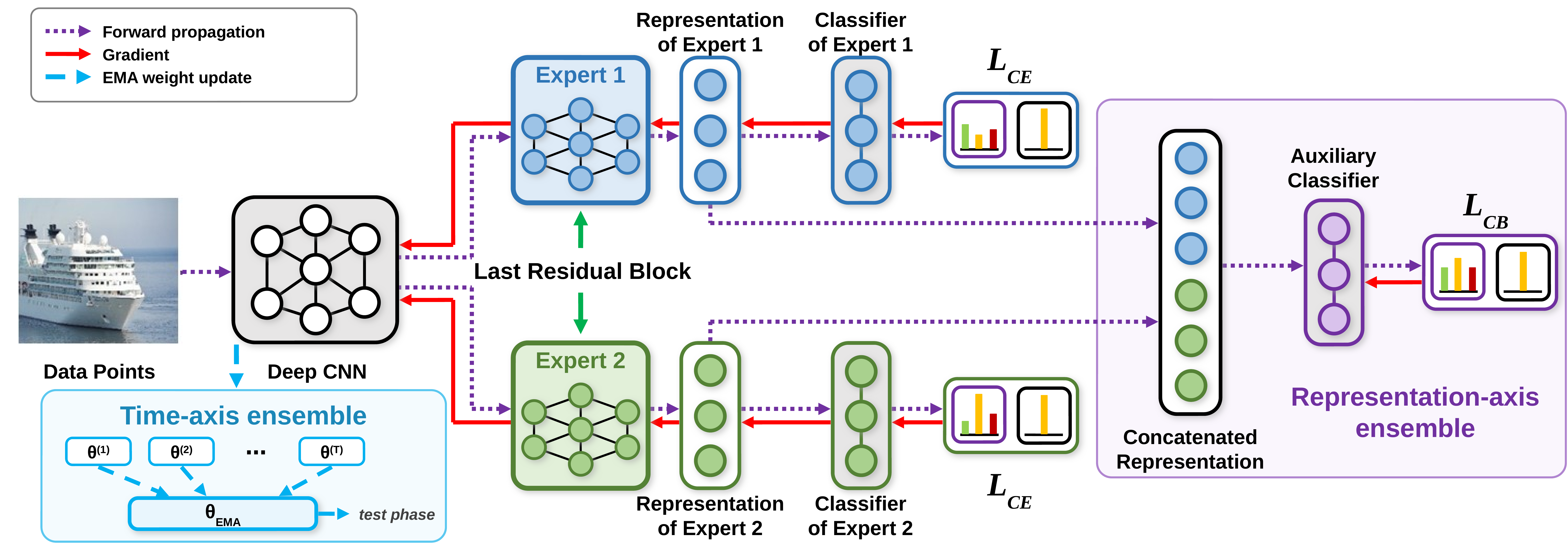}

	\caption{Overall structure of DAMEL (two experts). Purple lines indicate forward propagation, red lines indicate gradients, and the blue line indicates the updating of EMA weights at the end of each epoch. An auxiliary classifier with EMA weights is used in the test phase.}
	\label{structure}
\end{figure*}
By training multiple experts \textcolor{black}{independently}, DAMEL \textcolor{black}{encourages the learning of diverse} representations. Specifically, DAMEL \textcolor{black}{employs} $K$ experts that share a backbone $g(\cdot)$ and have individual \textcolor{black}{final} residual blocks $h_k(\cdot)$\textcolor{black}{, each followed by a} single-layer \textcolor{black}{classifier with} weights $w_k$, \textcolor{black}{where} $k\in\left\lbrace1,\ldots,K\right\rbrace$. We denote the output of the $k$th expert before its classification layer as $h_k\circ g(x_b)=z_k$, where $z_k$ \textcolor{black}{denotes} the \textcolor{black}{representation learned by} the $k$th \textcolor{black}{expert}. \textcolor{black}{The} prediction of the $k$th expert\textcolor{black}{,} $p_k(y|x_b)$\textcolor{black}{, is then computed} as $softmax(\alpha\overline{w}_k^T\overline{z}_k)$, where $\overline{w}_k^T$ and $\overline{z}_k$ denote the L2\textcolor{black}{-}normalized versions of $w_k^T$ and $z_k$, respectively, and $\alpha$ is a \textcolor{black}{scaling} hyperparameter. \textcolor{black}{Each} expert \textcolor{black}{is trained independently using} the cross-entropy loss:
\begin{equation}
\label{eq1}
	L_{CE}^{k}=\sum_{b=1}^{B}CELoss\left(y_{b},p_{k}\left(y|x_{b}\right)\right),
\end{equation}
where $CELoss$ denotes the cross-entropy loss. \textcolor{black}{Although all experts are} trained \textcolor{black}{using} the same loss \textcolor{black}{function}, \textcolor{black}{they learn} different representations because neural \textcolor{black}{networks initialized} with different \textcolor{black}{weights can converge to} different \textcolor{black}{modes during} optimization \cite{lakshminarayanan2017simple,nam2021diversity}. \textcolor{black}{From the perspective} of exploring multiple modes, training multiple experts \textcolor{black}{is conceptually related} to Bayesian neural networks \cite{blundell2015weight,wen2018flipout}\textcolor{black}{. However}, \textcolor{black}{deep ensemble methods have been shown to explore a broader range of modes and can therefore be} more effective than Bayesian neural networks \cite{fort2019deep}. In Section~\ref{analysis}, we \textcolor{black}{empirically} verify that \textcolor{black}{individual experts learn distinct} representations.
\subsection{Training the Auxiliary Classifier}
\label{trainingauxiliary}

DAMEL reduces both the squared bias and variance of predictions \textcolor{black}{by leveraging diverse} representations learned by multiple experts. Specifically, \textcolor{black}{the} representations \textcolor{black}{produced by the experts} are first concatenated\textcolor{black}{,} and an auxiliary classifier is \textcolor{black}{then} trained \textcolor{black}{on} the concatenated \textcolor{black}{representation using} a class-balanced loss. \textcolor{black}{We} aggregate the representations $\overline{z}_k$, $k=1,\ldots,K$\textcolor{black}{, through} concatenation \textcolor{black}{rather than} averaging to preserve their diversity.
Using the concatenated \textcolor{black}{representation}, the prediction \textcolor{black}{of} the auxiliary classifier\textcolor{black}{,} $p(y|x_b)$\textcolor{black}{, is computed} as $softmax(\alpha\overline{w}^T\overline{z})$, where $\overline{w}$ denotes the L2\textcolor{black}{-}normalized weights of the auxiliary classifier and $\overline{z}$ denotes the concatenated representations $[\overline{z}_1,\ldots,\overline{z}_K]$. \textcolor{black}{The} auxiliary classifier \textcolor{black}{is trained using} a class-balanced loss:
\begin{equation}
\label{eq2}
	L_{CB}=\sum_{b=1}^{B}CBLoss\left(y_{b},p\left(y|x_{b}\right)\right),
\end{equation}
where $CBLoss$ reweights the cross-entropy loss for each class \textcolor{black}{using} a factor inversely \textcolor{black}{related} to the \textcolor{black}{number} of \textcolor{black}{samples in that} class. \textcolor{black}{Recent studies have shown} that training \textcolor{black}{neural} networks with class-balanced losses \textcolor{black}{can lead to lower-quality learned} representations than training \textcolor{black}{them} with \textcolor{black}{the standard} cross-entropy loss \cite{kang2019decoupling}. \textcolor{black}{Accordingly}, we \textcolor{black}{apply gradients from} $CBLoss$ only \textcolor{black}{to} the auxiliary classifier and \textcolor{black}{stop} the gradients \textcolor{black}{from propagating back through} the concatenated \textcolor{black}{expert} representations.

Unlike existing multi-expert learning algorithms, DAMEL aggregates \textcolor{black}{the} representations, \textcolor{black}{rather than the} predictions, of multiple experts and \textcolor{black}{generates} predictions using \textcolor{black}{an} auxiliary classifier. This \textcolor{black}{design enables} the auxiliary classifier to \textcolor{black}{exploit} information from the representations \textcolor{black}{learned by} all experts. \textcolor{black}{By leveraging this} richer \textcolor{black}{and more diverse} information, the auxiliary classifier can produce more accurate predictions, \textcolor{black}{thereby reducing} the squared \textcolor{black}{prediction} bias. \textcolor{black}{In} contrast, if multiple auxiliary classifiers \textcolor{black}{are trained separately} and their predictions \textcolor{black}{are aggregated,} as in existing \textcolor{black}{approaches}, each classifier \textcolor{black}{can access only} the \textcolor{black}{representation learned by} a single expert. In Section~\ref{analysis}, we \textcolor{black}{demonstrate} that aggregating the representations of multiple experts \textcolor{black}{outperforms prediction aggregation} by reducing \textcolor{black}{both} the squared bias \textcolor{black}{and} the variance of predictions.
\subsection{End-to-End Training of DAMEL}
\label{endtoend}
Unlike the recent trend \textcolor{black}{in CIL toward decoupling representation learning from the training} of class-balanced classifiers \cite{kang2019decoupling}, we train the auxiliary classifier and multiple experts \textcolor{black}{in} DAMEL end-to-end \textcolor{black}{using} the \textcolor{black}{following} total loss\textcolor{black}{,} $L_{total}$:
\begin{equation}
\label{eq3}
	L_{total}=L_{CB}+\sum_{k=1}^{K}L_{CE}^{k}.
\end{equation}
End-to-end training \textcolor{black}{enables} the auxiliary classifier to be \textcolor{black}{jointly optimized} with the concatenated \textcolor{black}{expert} representations \cite{lee2021abc,wang2021contrastive}. In Section~\ref{ablation}, we \textcolor{black}{demonstrate} that end-to-end training of DAMEL \textcolor{black}{achieves} better performance than decoupled training of \textcolor{black}{the} multiple experts and the auxiliary classifier.

\textcolor{black}{During testing}, we use \textcolor{black}{only} the predictions of the auxiliary classifier \textcolor{black}{and exclude those of} the \textcolor{black}{individual} experts because \textcolor{black}{the experts} are not trained with \textcolor{black}{a} class-balanced loss. \textcolor{black}{The overall architecture} of DAMEL is presented in Section~\ref{structure}.
\subsection{EMA \textcolor{black}{Weight Aggregation}}
\label{ema}
To \textcolor{black}{obtain an} ensemble effect \textcolor{black}{from} multiple experts along the time axis, we \textcolor{black}{additionally} aggregate the network weights at the end of each epoch\textcolor{black}{. This reduces} variability \textcolor{black}{in} the network \textcolor{black}{parameters} and, consequently, the variance of the predictions. Let $\boldsymbol{\theta}$ denote \textcolor{black}{all network} weights trained \textcolor{black}{using} the total loss in Equation~\eqref{eq3}. We \textcolor{black}{define} the EMA weights\textcolor{black}{,} $\boldsymbol{\theta}_{EMA}$\textcolor{black}{,} by aggregating $\boldsymbol{\theta}$ at the end of each epoch as follows:
\begin{equation}
\label{eq4}
	\boldsymbol{\theta}_{EMA}^{t}=\left(1-\beta\right)\times \boldsymbol{\theta}_{EMA}^{t-1}+\beta \times\boldsymbol{\theta},
\end{equation}
where $\boldsymbol{\theta}_{EMA}^{t}$ denotes the EMA weights at the $t$th epoch and $\beta$ is \textcolor{black}{the} EMA update hyperparameter. The \textcolor{black}{network} weights $\boldsymbol{\theta}$ \textcolor{black}{are trained independently} of the aggregated weights $\boldsymbol{\theta}_{EMA}$, and DAMEL \textcolor{black}{uses} only $\boldsymbol{\theta}_{EMA}$ \textcolor{black}{during testing}. As shown in Equation~\eqref{eq4}, $\boldsymbol{\theta}_{EMA}$ is not \textcolor{black}{updated through backpropagation but is instead updated automatically using the} trained \textcolor{black}{network weights} $\boldsymbol{\theta}$. \textcolor{black}{Consequently}, the \textcolor{black}{additional computational} complexity \textcolor{black}{introduced by} EMA \textcolor{black}{weight aggregation} is \textcolor{black}{minimal}. The only additional \textcolor{black}{step} is to compute the running mean and \textcolor{black}{variance for} each \textcolor{black}{network} layer \textcolor{black}{using} $\boldsymbol{\theta}_{EMA}$, \textcolor{black}{as} required for batch normalization \textcolor{black}{during testing}. These statistics can be \textcolor{black}{readily obtained} by \textcolor{black}{performing a forward pass over} the training set before the test phase.

Using \textcolor{black}{an} ensemble along the additional \textcolor{black}{time} axis, \textcolor{black}{DAMEL} can further \textcolor{black}{reduce prediction} variance compared \textcolor{black}{with} existing multi-expert learning algorithms. It is \textcolor{black}{well established} that the variance of \textcolor{black}{ensemble} predictions decreases as the correlation among the component predictions decreases. \textcolor{black}{Accordingly}, we update $\boldsymbol{\theta}_{EMA}$ \textcolor{black}{at the end of} each epoch, rather than \textcolor{black}{after every} iteration, to increase the diversity of the \textcolor{black}{network} weights \textcolor{black}{included in} the ensemble. 

In addition to reducing \textcolor{black}{prediction} variance, EMA \textcolor{black}{weight aggregation} also reduces the squared bias of predictions because the aggregated weights stably approach wide local optima \cite{izmailov2018averaging}, which \textcolor{black}{remain near-}optimal \textcolor{black}{under} small perturbations. This stable optimization process \textcolor{black}{enables the use of} a \textcolor{black}{relatively} high learning rate, \textcolor{black}{facilitating} convergence of the test loss to a low value, as \textcolor{black}{analyzed} in Appendix~A. \textcolor{black}{In Appendix}~\textcolor{black}{B, we} present confusion matrices \textcolor{black}{for} predictions on the CIFAR-10 \textcolor{black}{test set} \cite{krizhevsky2009learning} to show that the use of EMA weights reduces the squared \textcolor{black}{prediction} bias for minority classes.


EMA \textcolor{black}{weight aggregation} can \textcolor{black}{also serve as} a \textcolor{black}{general} framework \textcolor{black}{applicable to} CIL \textcolor{black}{and} other \textcolor{black}{learning} tasks. \textcolor{black}{The} pseudocode for the \textcolor{black}{DAMEL} training procedure \textcolor{black}{is presented} in \textcolor{black}{Algorithm}~\ref{alg}. The for-loops in the algorithm can be executed in parallel.

\vspace{-0.15in}
\begin{algorithm}[H]
   \caption{Training procedure of DAMEL}
   \label{alg}
\begin{algorithmic}
   \STATE {\bfseries Input:} $\mathcal{MB} = \left\lbrace\left(x_{b},y_{b}\right):  b\in\left(1,\ldots,B\right)\right\rbrace \subset \mathcal{D}$
   \STATE {\bfseries Output:} Weights of the network $\boldsymbol{\theta}$, $\boldsymbol{\theta}_{EMA}$
   \WHILE{training}
    
   \FOR{$b=1$ to $B$}
   \FOR{$k=1$ to $K$}
   \STATE $z_{k}=h_{k}\circ g\left(x_{b}\right)$
   \STATE $p_{k}\left(y|x_{b}\right)=softmax\left(\alpha\overline{w}^{T}_{k}\overline{z}_{k}\right)$
   \STATE {\bfseries Calculate} $L_{CE}^{k}+=CELoss\left(y,p_{k}\left(y_{b}|x_{b}\right)\right)$
   \ENDFOR
   \STATE Concatenated representations $z=[z_{1},\ldots,z_{k}]$
   \STATE $p\left(y|x_{b}\right)=softmax\left(\alpha\overline{w}^{T}\overline{z}\right)$
   \STATE {\bfseries Calculate} $L_{CB}+=CBLoss\left(y,p\left(y_{b}|x_{b}\right)\right)$
   \ENDFOR
   \STATE Total loss $L_{total}=L_{CB}+\sum_{k=1}^{K}L_{CE}^{k}$
   \STATE $\Delta\boldsymbol{\theta}\propto\nabla_{\boldsymbol{{\theta}}}L_{total}$, $\quad\boldsymbol{\theta}\gets\boldsymbol{\theta}+\Delta\boldsymbol{\theta}$
   \IF {Epoch ends}
   \STATE $\boldsymbol{\theta}_{EMA}^{t}=\left(1-\beta\right)\times\boldsymbol{\theta}_{EMA}^{t-1}+\beta \times\boldsymbol{\theta}$
   \ENDIF
   \ENDWHILE
\end{algorithmic}
\end{algorithm}

\section{Experiments}
\label{experiments}

\subsection{Experimental Setup}

\label{setup}
\textbf{CIFAR-10-LT and CIFAR-100-LT} \cite{cui2019class} are artificially \textcolor{black}{constructed} long-tailed datasets \textcolor{black}{derived} from CIFAR-10 and CIFAR-100 \cite{krizhevsky2009learning}, respectively\textcolor{black}{. The} number of \textcolor{black}{training samples decreases} exponentially from $N_{1}$ to $N_{L}$, \textcolor{black}{such that} $N_{k}=N_{1}\times\left(N_{L}/N_{1}\right)^{\frac{k-1}{L-1}}$. \textcolor{black}{Each image has a resolution} of 32 $\times$ 32~\textcolor{black}{pixels}, and \textcolor{black}{data augmentation was performed using} random \textcolor{black}{cropping} and horizontal \textcolor{black}{flipping}. We conducted experiments on CIFAR-10-LT and CIFAR-100-LT \cite{cui2019class} \textcolor{black}{with} class imbalance \textcolor{black}{ratios} $\gamma$ \textcolor{black}{of} $100$ to $50$. For CIFAR-100-LT, we categorized the classes into three groups: Many (classes with more than 100 training \textcolor{black}{samples}), Medium (classes with 20--100 training \textcolor{black}{samples}), and Few (classes with \textcolor{black}{fewer} than 20 \textcolor{black}{training samples}), following \cite{liu2019large}. We \textcolor{black}{evaluated both} group-wise and overall \textcolor{black}{classification} accuracy\textcolor{black}{, consistent with} baseline algorithms \textcolor{black}{such as} BBN, LFME, ACE, RIDE, \textcolor{black}{and} TLC. \textcolor{black}{Although} metrics such as precision\textcolor{black}{,} recall\textcolor{black}{,} F-1 \textcolor{black}{score, and} AUC are \textcolor{black}{commonly} reported \textcolor{black}{for} binary \textcolor{black}{CIL}, fine-grained accuracy \textcolor{black}{across the Many, Medium, and Few groups} is \textcolor{black}{widely used} in multi-class \textcolor{black}{CIL} settings. \textcolor{black}{We} repeated the experiments on the CIFAR datasets five times \textcolor{black}{using different} random seeds and \textcolor{black}{reported} the average accuracy \textcolor{black}{across runs}. \textcolor{black}{Consistent with previous} CIL studies, we \textcolor{black}{used} ResNet-32 \cite{he2016deep} \textcolor{black}{as the backbone network}. 

\textbf{ImageNet-LT} \cite{liu2019large} is a large-scale long-tailed dataset \textcolor{black}{derived} from ImageNet-2012 \cite{deng2009imagenet}\textcolor{black}{. It contains} 1000 classes \textcolor{black}{whose sample frequencies follow} a Pareto distribution\textcolor{black}{,} with $N_{1}=1280$ \textcolor{black}{samples in the largest class} and $N_{1000}=5$ \textcolor{black}{samples in the smallest class}. Each image was resized to 224 $\times$ 224~\textcolor{black}{pixels}, and \textcolor{black}{data augmentation was performed using} random \textcolor{black}{cropping}, horizontal \textcolor{black}{flipping}, and color \textcolor{black}{jittering}. Following previous studies, we conducted experiments on ImageNet-LT \textcolor{black}{using} ResNet-10, ResNet-50, and ResNeXt-50 \cite{xie2017aggregated} \textcolor{black}{as the backbone networks}.

\textbf{iNaturalist2018} \cite{van2018inaturalist} is a large-scale dataset comprising real-world \textcolor{black}{images of} animals and plants\textcolor{black}{. It contains} 8142 classes \textcolor{black}{with a highly} imbalanced \textcolor{black}{class} distribution\textcolor{black}{, ranging from} $N_{1}=1000$ \textcolor{black}{samples in the largest class to} $N_{8142}=2$ \textcolor{black}{samples in the smallest class}. Each image was resized to 224 $\times$ 224~\textcolor{black}{pixels}, and \textcolor{black}{data augmentation was performed using} random \textcolor{black}{cropping} and horizontal \textcolor{black}{flipping}. \textcolor{black}{Consistent with previous} CIL studies, we conducted experiments on iNaturalist2018 \textcolor{black}{using} ResNet-50 \textcolor{black}{as the backbone network}.   

We \textcolor{black}{evaluated} prediction accuracy every 10 epochs on \textcolor{black}{the} CIFAR-10 and CIFAR-100 \textcolor{black}{test sets} and every 20 epochs on \textcolor{black}{the} ImageNet-LT and iNaturalist2018 \textcolor{black}{test sets}. For all the datasets, \textcolor{black}{DAMEL was} trained for 200 epochs using stochastic gradient descent with a learning rate of 0.1 and a momentum of 0.9, \textcolor{black}{following a setup} similar to \textcolor{black}{that used in} the \textcolor{black}{ACE} study \cite{cai2021ace}. The EMA update hyperparameter $\beta$ was set to 0.1. \textcolor{black}{For} ImageNet-LT and iNaturalist2018\textcolor{black}{, we report the performance of DAMEL using three and four experts}. \textcolor{black}{Additional} details regarding the hyperparameter settings and implementation are \textcolor{black}{provided} in Appendix~C. We compared DAMEL with the baseline \textcolor{black}{methods evaluated} in ACE \cite{cai2021ace} and RIDE \cite{wang2021longtailed}. Specifically, \textcolor{black}{the comparison included the following categories: (}1\textcolor{black}{) a} vanilla \textcolor{black}{method, consisting of a standard} deep CNN trained \textcolor{black}{with} cross-entropy loss\textcolor{black}{; (}2\textcolor{black}{)} CIL \textcolor{black}{methods, including} OLTR \cite{liu2019large}, \textcolor{black}{resampling} \cite{chawla2002smote}, M2m \cite{kim2020m2m}, Mixup \cite{zhang2018mixup}, LDAM \cite{NEURIPS2019_621461af}, Logit \textcolor{black}{Adjustment} \cite{menon2020long}, BALMS \cite{NEURIPS2020_2ba61cc3}, cRT, LWS, $\tau$-norm \cite{kang2019decoupling}, PACO \cite{cui2021parametric}, TSC \cite{li2022targeted}, ProCo \cite{du2024proco}, Meta-CALA \cite{zhou2025cala}, and CSA \cite{shi2023resampling}\textcolor{black}{;} and \textcolor{black}{(}3\textcolor{black}{)} multi-expert learning \textcolor{black}{methods, including} BBN \cite{zhou2020bbn}, LFME \cite{xiang2020learning}, RIDE \cite{wang2021longtailed}, ACE \cite{cai2021ace}, TLC \cite{li2022trustworthy}, ResLT \cite{cui2022reslt}, and MCE \cite{wang2024mce}. Further details of the baseline \textcolor{black}{methods} are provided in Appendix~D.
\subsection{Experimental Results}
\label{results}
The prediction \textcolor{black}{accuracies} of the baseline algorithms and DAMEL trained on CIFAR-100-LT with $\gamma=$100 are summarized in Table \ref{cifar100}. \textcolor{black}{Notably,} even with \textcolor{black}{a single} expert\textcolor{black}{,} without \textcolor{black}{representation aggregation}, the proposed \textcolor{black}{method} achieved \textcolor{black}{the} best overall performance. \textcolor{black}{This result demonstrates} the effectiveness of aggregating \textcolor{black}{network} weights along the time axis. \textcolor{black}{When} the representations of multiple experts \textcolor{black}{were aggregated}, DAMEL \textcolor{black}{achieved further} performance \textcolor{black}{gains} and \textcolor{black}{obtained} state-of-the-art \textcolor{black}{results across} the Many, Medium, and Few \textcolor{black}{class groups}. We also \textcolor{black}{observed} that the performance of DAMEL consistently improved as the number of experts increased. In Section \ref{analysis}, we \textcolor{black}{show} that the superior performance of DAMEL \textcolor{black}{results} from \textcolor{black}{a} reduction \textcolor{black}{in both} squared bias and \textcolor{black}{prediction} variance \textcolor{black}{through} aggregation along the two axes.
\begin{table}[htbp]
  \caption{Prediction accuracy \textcolor{black}{(\%)} of baseline algorithms and DAMEL. The results were copied from ACE \cite{cai2021ace} and RIDE \cite{wang2021longtailed}. ``-" indicate\textcolor{black}{s} that the corresponding results are not reported in the paper. Algorithms with * aggregate the predictions of four experts.}
  \centering
  \resizebox{4.25in}{!}{%
  \begin{tabular}{c|c|c|c|c}    \toprule
    \multicolumn{5}{c}{\textbf{CIFAR-100-LT with $\gamma=$100}}                   \\
    \midrule
    \textbf{Algorithm}   &\textbf{Overall}&\textbf{Many}&\textbf{Medium}&\textbf{Few}\\
    \midrule
    Vanilla &$38.3$&$65.2$&$37.1$&$9.1$ \\
    Resampling \cite{chawla2002smote}&$36.0$&$59.0$&$35.4$&$10.9$ \\
    Mixup \cite{zhang2018mixup}&$41.2$&$70.7$&$40.4$&$8.8$ \\
    LDAM+DRW \cite{NEURIPS2019_621461af}&$42.0$&$61.5$&$41.7$&$20.2$ \\
    LDAM+M2m \cite{kim2020m2m}&$43.5$&-&-&- \\
    Logit Adj \cite{menon2020long}&$43.9$&-&-&- \\
    cRT \cite{kang2019decoupling}&$43.3$&$64.0$&$44.8$&$18.1$ \\
    \midrule
    BBN \cite{zhou2020bbn}&$39.4$&$47.2$&$49.4$&$19.8$ \\
    LFME \cite{xiang2020learning}&$43.8$&-&-&- \\
    RIDE* \cite{wang2021longtailed}&$49.1$&$69.3$&$49.3$&$26.0$ \\
    ACE* \cite{cai2021ace}&$49.6$&$66.3$&$52.8$&$27.2$ \\
    TLC* \cite{li2022trustworthy}&$49.8$&$71.1$&$48.4$&$29.7$ \\
    \midrule
    \textbf{DAMEL (1 expert)}&$\mathbf{50.0}$&$\mathbf{66.5}$&$\mathbf{51.0}$&$\mathbf{30.3}$ \\
    \textbf{DAMEL (2 experts)}&$\mathbf{52.1}$&$\mathbf{69.5}$&$\mathbf{53.1}$&$\mathbf{31.2}$ \\
    \textbf{DAMEL (3 experts)}&$\mathbf{52.6}$&$\mathbf{70.4}$&$\mathbf{53.5}$&$\mathbf{31.5}$ \\
    \textbf{DAMEL (4 experts)}&$\mathbf{53.1}$&$\mathbf{71.6}$&$\mathbf{53.8}$&$\mathbf{31.4}$ \\
    \bottomrule

  \end{tabular}}

 \label{cifar100}
\end{table}

Moreover, to evaluate DAMEL under \textcolor{black}{different class-imbalance} settings, we conducted experiments on both CIFAR-10-LT and CIFAR-100-LT \textcolor{black}{with class-imbalance} $\gamma$ \textcolor{black}{ratios of} 50 \textcolor{black}{and} 100. \textcolor{black}{As shown in} Table \ref{cifar10}, DAMEL with \textcolor{black}{four} experts \textcolor{black}{outperformed} the baseline algorithms under \textcolor{black}{all evaluated settings}. 

\begin{table}[htbp] 
  \caption{Prediction accuracy \textcolor{black}{(\%)} of baseline algorithms and DAMEL\textcolor{black}{.}}
  \centering
  \resizebox{4in}{!}{%
  \begin{tabular}{c|c|c|c|c}    \toprule
   &\multicolumn{2}{c|}{\textbf{CIFAR-10-LT}}&\multicolumn{2}{c}{\textbf{CIFAR-100-LT}} \\
    \midrule
        \textbf{Algorithm}&\textbf{$\gamma=$100}&$\,$\textbf{$\gamma=$50}$\,$&\textbf{$\gamma=$100}&$\,$\textbf{$\gamma=$50}$\,$\\
    \midrule
    Vanilla &$69.8$&$75.2$&$38.3$&$42.1$ \\
    Mixup \cite{zhang2018mixup}&$73.1$&$77.8$&$41.2$&$45.0$ \\
    LDAM+DRW \cite{NEURIPS2019_621461af}&$77.0$&$79.3$&$42.0$&$45.1$ \\
    LDAM+M2m \cite{kim2020m2m}&$79.1$&-&$43.5$&- \\
    Logit Adj \cite{menon2020long}&$77.7$&-&$43.9$&- \\
    TSC \cite{li2022targeted}&$79.7$&$82.9$&$43.8$&$47.4$ \\
    \textcolor{black}{ProCo \cite{du2024proco}} &\textcolor{black}{84.8} & \textcolor{black}{87.5} & \textcolor{black}{51.8} & \textcolor{black}{56.4} \\
    \textcolor{black}{Meta-CALA \cite{zhou2025cala}}&\textcolor{black}{84.8} & \textcolor{black}{-} &\textcolor{black}{52.3} &  \textcolor{black}{-}  \\
    \textcolor{black}{CSA \cite{shi2023resampling}}& \textcolor{black}{82.5} & \textcolor{black}{86.0} & \textcolor{black}{46.6} & \textcolor{black}{51.9}  \\
    \midrule
    BBN \cite{zhou2020bbn}&$79.8$&$82.2$&$39.4$&$47.0$ \\
    RIDE \cite{wang2021longtailed}&$81.7$&$84.0$&$48.0$&$52.2$ \\
    ACE \cite{cai2021ace}&$81.2$&$84.3$&$49.4$&$50.7$ \\
    TLC \cite{li2022trustworthy}&$80.3$&-&$49.0$&- \\
    ResLT \cite{cui2022reslt}&-&$85.4$&-&$52.8$ \\
    \textcolor{black}{MCE \cite{wang2024mce}} &\textcolor{black}{84.5} & \textcolor{black}{87.3} & \textcolor{black}{50.1} & \textcolor{black}{56.0}  \\
    \midrule
    \textbf{DAMEL (3 experts)}&$\mathbf{84.7}$&$\mathbf{86.5}$&$\mathbf{52.6}$&$\mathbf{56.9}$ \\ 
    \textcolor{black}{\textbf{DAMEL (4 experts)}}&\textcolor{black}{\textbf{85.0}}&\textcolor{black}{\textbf{87.9}}&\textcolor{black}{\textbf{53.1}}&\textcolor{black}{\textbf{57.8}}\\
    \bottomrule
  \end{tabular}}
\label{cifar10}
\end{table}

\textcolor{black}{The} DAMEL and baseline \textcolor{black}{models reported} in Table \ref{cifar100} and \ref{cifar10} were trained \textcolor{black}{using} weak data augmentation techniques\textcolor{black}{, including} random \textcolor{black}{cropping} and random horizontal \textcolor{black}{flipping}. To \textcolor{black}{assess} DAMEL \textcolor{black}{against} recent \textcolor{black}{methods} that \textcolor{black}{employ stronger} data augmentation, we \textcolor{black}{conducted} additional experiments and \textcolor{black}{report} the results in Table \ref{strong}. \textcolor{black}{Specifically}, we \textcolor{black}{compared DAMEL with} BALMS \cite{NEURIPS2020_2ba61cc3} and PACO \cite{cui2021parametric}, which use AutoAugment \cite{cubuk2019autoaugment} and Cutout \cite{devries2017improved} as strong data augmentation techniques. For \textcolor{black}{a} fair comparison, we also \textcolor{black}{applied} AutoAugment and Cutout when training DAMEL. \textcolor{black}{As shown in Table}~\ref{strong}, DAMEL achieved higher performance than \textcolor{black}{both} BALMS and PACO. These results \textcolor{black}{indicate} that DAMEL can be \textcolor{black}{effectively} combined with advanced data augmentation techniques. \textcolor{black}{Further details on} AutoAugment and Cutout \textcolor{black}{are provided} in Appendix E.

\begin{table}[htbp]
  \caption{Prediction accuracy \textcolor{black}{(\%)} of algorithms trained with AutoAugment and Cutout.}
  \centering
  \resizebox{3.55in}{!}{%
  \begin{tabular}{c|c|c|c}    \toprule
    \multicolumn{4}{c}{\textbf{CIFAR-100-LT}} \\
    \midrule
       \textbf{Algorithm}&\textbf{$\gamma=$100}&$\,$\textbf{$\gamma=$50}$\,$&$\,\,$\textbf{$\gamma=$10}$\,\,$\\
    \midrule
    BALMS \cite{NEURIPS2020_2ba61cc3}&$50.8$&$54.2$&$63.0$\\
    PACO \cite{cui2021parametric}&$52.0$&$56.0$&$64.2$ \\
    \midrule
    \textbf{DAMEL (3 experts)}&$\mathbf{54.5}$&$\mathbf{58.3}$&$\mathbf{66.6}$\\
    \bottomrule
  \end{tabular}}
\label{strong}
\end{table}


Furthermore, we evaluated DAMEL on the large-scale ImageNet-LT and iNaturalist2018 \textcolor{black}{datasets}. \textcolor{black}{As shown in} Table \ref{imagenet}, DAMEL with four experts consistently outperformed the baseline algorithms. These results \textcolor{black}{demonstrate} that DAMEL \textcolor{black}{remains effective} even on large-scale \textcolor{black}{long-tailed} datasets.

\begin{table}[H]
\caption{Prediction accuracy \textcolor{black}{(\%)} of baseline algorithms and DAMEL\textcolor{black}{.}}
  \centering
  \resizebox{4.5in}{!}{%
  \begin{tabular}{c|c|c|c|c}    \toprule
    &\multicolumn{3}{c|}{\textbf{ImageNet-LT}}&\textbf{iNaturalist} \\
\midrule
    \textbf{Algorithm}   &$\,\;$\textbf{Res10}$\,\;$&$\,\,$\textbf{Res50}$\,\,$&\textbf{ResX50}&\textbf{Res50}\\
    \midrule
    Vanilla &$20.9$&$41.6$&$44.4$&$66.1$ \\
    OLTR \cite{liu2019large}&$34.1$&-&$46.3$&$63.9$ \\
    LDAM+DRW \cite{NEURIPS2019_621461af}&$36.0$&-&-&$68.0$ \\
    $\tau$-norm \cite{kang2019decoupling}&$40.6$&$46.7$&$49.4$&$65.6$ \\
    cRT \cite{kang2019decoupling}&$41.8$&$47.3$&$49.5$&$65.2$ \\
    LWS \cite{kang2019decoupling}&$41.4$&$47.7$&$49.9$&$65.9$ \\
    Logit Adj \cite{menon2020long}&-&$51.1$&-&$66.4$ \\
    TSC \cite{li2022targeted}&-&$52.4$&-&$69.7$\\
    \textcolor{black}{ProCo \cite{du2024proco}} & \textcolor{black}{-}  & \textcolor{black}{57.3} & \textcolor{black}{58.0} & \textcolor{black}{73.5} \\
    \textcolor{black}{Meta-CALA \cite{zhou2025cala}}&\textcolor{black}{-} &\textcolor{black}{-} & \textcolor{black}{55.1} & \textcolor{black}{74.1}\\
    \textcolor{black}{CSA \cite{shi2023resampling}}&\textcolor{black}{ - }& \textcolor{black}{49.7} &\textcolor{black}{ - }&\textcolor{black}{ -} \\
    \midrule
    BBN \cite{zhou2020bbn}&-&$48.3$&$49.3$&$68.0$ \\
    LFME \cite{xiang2020learning}&$38.8$&-&-&- \\
    RIDE \cite{wang2021longtailed}&$45.9$&$54.9$&$56.4$&$72.2$ \\
    ACE \cite{cai2021ace}&$44.0$&$54.7$&$56.6$&$72.9$ \\
    TLC \cite{li2022trustworthy}&-&$54.6$&-&- \\
    ResLT \cite{cui2022reslt}&$43.8$&$52.9$&-&$70.5$ \\
    \textcolor{black}{MCE \cite{wang2024mce}} & \textcolor{black}{-}&\textcolor{black}{ -} & \textcolor{black}{58.8} & \textcolor{black}{73.2} \\
    \midrule
    \textbf{DAMEL (3 experts)}&$\mathbf{47.7}$&$\mathbf{56.8}$&$\mathbf{58.3}$&$\mathbf{73.6}$\\
    \textcolor{black}{\textbf{DAMEL (4 experts)}}&\textcolor{black}{\textbf{48.8}}&\textcolor{black}{\textbf{57.7}}&\textcolor{black}{\textbf{59.5}}&\textcolor{black}{\textbf{74.4}}\\
    \bottomrule
  \end{tabular}}

\label{imagenet}
\end{table}

\subsection{Detailed Analysis of DAMEL}
\label{analysis}

\begin{figure}[ht]
	\begin{center}
		\begin{tabular}{cccc}
			\includegraphics[width=0.2\textwidth]{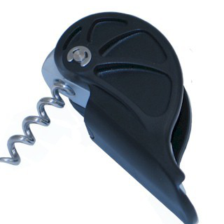}  &  \includegraphics[width=0.2\textwidth]{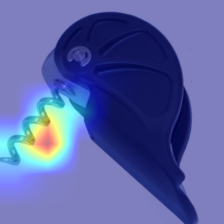}&\includegraphics[width=0.2\textwidth]{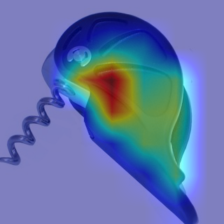}  &   \includegraphics[width=0.2\textwidth]{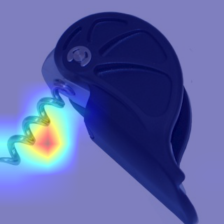} \\
			
			\footnotesize{(a) Corkscrew } & \footnotesize{(b) Expert 1 } &
			\footnotesize{(c) Expert 2 } & \footnotesize{(d) Expert 3} \\
		\end{tabular}
	\end{center}
	\caption{
		Image representations taken differently by multiple experts.}
	\label{differentrep}
\end{figure}

Based on the \textcolor{black}{observation} that neural \textcolor{black}{networks} with different parameter initializations \textcolor{black}{can converge to} different \textcolor{black}{modes during} optimization \cite{lakshminarayanan2017simple}, we \textcolor{black}{hypothesize} that multiple experts learn \textcolor{black}{distinct} representations even \textcolor{black}{when} trained with the same loss \textcolor{black}{function}. To verify this \textcolor{black}{hypothesis}, we compared visualizations of dimension-reduced features, particularly principal components \cite{lee2013dependence,soh2018application,chung2019crime}, derived from \textcolor{black}{the} representations learned \textcolor{black}{by} multiple experts using EigenCAM \cite{muhammad2020eigen,kho2022exploiting}. \textcolor{black}{As shown in} Figure \ref{differentrep} (b) and Figure \ref{differentrep} (d), the first and third experts learned representations corresponding to the screw \textcolor{black}{portion} of the corkscrew. Using these representations, \textcolor{black}{both} experts correctly classified the image in Figure \ref{differentrep} (a) as \textcolor{black}{a} corkscrew. \textcolor{black}{In} contrast, the second expert learned \textcolor{black}{a} different \textcolor{black}{representation focused on} the handle of the corkscrew and incorrectly classified the image as a neck brace. Nevertheless, the auxiliary classifier \textcolor{black}{produced the} correct prediction because \textcolor{black}{it received informative} representations \textcolor{black}{from} the first and third experts. \textcolor{black}{This} example \textcolor{black}{illustrates that} aggregating representations from multiple experts \textcolor{black}{increases the likelihood that} the auxiliary classifier \textcolor{black}{receives} appropriate \textcolor{black}{discriminative information}, even when some experts learn \textcolor{black}{less informative or misleading} representations. \textcolor{black}{Therefore, in} addition to reducing \textcolor{black}{prediction} variance \textcolor{black}{through} the ensemble effect, DAMEL can \textcolor{black}{also} reduce squared \textcolor{black}{prediction} bias by \textcolor{black}{leveraging} appropriate representations \textcolor{black}{from multiple experts}. \textcolor{black}{Additional} examples of experts learning \textcolor{black}{distinct} representations \textcolor{black}{are provided} in Appendix F.

\textcolor{black}{We further investigated} whether aggregating the representations of multiple experts \textcolor{black}{yields} better performance than aggregating \textcolor{black}{their} predictions. \textcolor{black}{As shown in} Table \ref{repanalysis}, \textcolor{black}{representation aggregation} consistently \textcolor{black}{achieved} higher performance \textcolor{black}{than prediction aggregation}. \textcolor{black}{Moreover,} the performance gap between the two approaches increased as the number of experts increased. This \textcolor{black}{result} may be \textcolor{black}{attributed to the ability of} the auxiliary classifier \textcolor{black}{to more effectively exploit complementary} information \textcolor{black}{contained in} the representations learned \textcolor{black}{by} multiple experts, as \textcolor{black}{illustrated} in Figure \ref{differentrep}. These \textcolor{black}{findings} demonstrate that aggregating representations is more \textcolor{black}{effective} than aggregating predictions \textcolor{black}{for producing} accurate predictions \textcolor{black}{based on information} learned by all experts.

\begin{table}[htbp]
  \caption{Prediction accuracy \textcolor{black}{(\%)} of DAMEL using prediction aggregation versus representation aggregation on ImageNet-LT with ResNet-10.}
  \centering 
  \begin{tabular}{ccc}
    \toprule     
    \textbf{Number of experts}&\textbf{Aggregation of}&\textbf{Aggregation of}\\
    &\textbf{predictions}&\textbf{representations}\\
    \midrule
    With 2 experts&$45.8$&$\mathbf{46.6}$ \\
    With 3 experts&$46.5$&$\mathbf{47.7}$ \\
    With 4 experts&$46.9$&$\mathbf{48.8}$\\
    \bottomrule
  
  \end{tabular}

\label{repanalysis}
\end{table}

We also \textcolor{black}{hypothesized} that EMA weights stably approach local optima in the test \textcolor{black}{error landscape}. To verify this \textcolor{black}{hypothesis}, we \textcolor{black}{compared the} test errors \textcolor{black}{of} algorithms trained with and without using EMA \textcolor{black}{weight aggregation, as reported} in Appendix A. 

To \textcolor{black}{further examine whether} EMA \textcolor{black}{weight aggregation reduces} squared \textcolor{black}{prediction} bias \textcolor{black}{for} minority classes, we compared confusion matrices on the CIFAR-10-LT \textcolor{black}{test set for} the vanilla algorithm trained with cross-entropy loss, DAMEL without EMA weights, RIDE, and \textcolor{black}{the full} DAMEL \textcolor{black}{model}. The \textcolor{black}{corresponding} results are \textcolor{black}{presented} in Appendix B.

Furthermore, to \textcolor{black}{quantitatively assess} whether EMA \textcolor{black}{weight aggregation} and \textcolor{black}{representation concatenation reduce} both the squared bias and \textcolor{black}{prediction} variance, we analyzed \textcolor{black}{these quantities for widely used baseline algorithms, as summarized in Table}~\ref{biasvariancetable}\textcolor{black}{. To compute} squared bias and variance \textcolor{black}{for multiclass} classification, we \textcolor{black}{converted each prediction into a one-hot vector by assigning a value of 1 to} the class \textcolor{black}{with} the highest predicted probability and \textcolor{black}{0 to all remaining} classes, following previous studies. We \textcolor{black}{then estimated} squared bias and variance by repeating experiments for each algorithm 20 times \textcolor{black}{using different} random \textcolor{black}{seeds associated with} random sampling of \textcolor{black}{the} training set. \textcolor{black}{For} the vanilla algorithm, \textcolor{black}{prediction} variance was relatively low, \textcolor{black}{whereas} squared \textcolor{black}{prediction} bias was \textcolor{black}{substantially} higher than that of the other \textcolor{black}{methods} because the classifier was not trained with a class-balanced loss. \textcolor{black}{Consistent with} the \textcolor{black}{findings} of \cite{wang2021longtailed}, rebalancing \textcolor{black}{methods} such as LDAM+DRW \cite{NEURIPS2019_621461af} and cRT \cite{kang2019decoupling} reduced squared \textcolor{black}{prediction} bias but \textcolor{black}{increased prediction} variance \textcolor{black}{relative} to the vanilla algorithm. \textcolor{black}{By employing} multiple experts, RIDE \cite{wang2021longtailed} reduced \textcolor{black}{prediction} variance and consequently improved \textcolor{black}{classification} accuracy. We also \textcolor{black}{evaluated} the squared bias and variance of DAMEL without EMA weights and DAMEL without \textcolor{black}{representation aggregation} to \textcolor{black}{isolate} the \textcolor{black}{contribution} of the ensemble along each axis. \textcolor{black}{The results show} that \textcolor{black}{ensembling} along \textcolor{black}{either} axis reduces \textcolor{black}{both} squared bias and \textcolor{black}{prediction} variance compared \textcolor{black}{with methods} that do not \textcolor{black}{employ} multiple experts. \textcolor{black}{By jointly leveraging} multiple experts along \textcolor{black}{both} the \textcolor{black}{representation and time} axes, DAMEL further \textcolor{black}{reduces} both squared bias and \textcolor{black}{prediction} variance.
\begin{table}[htbp]
\caption{Analysis of the squared bias and variance of the predictions on CIFAR-100-LT with $\gamma$=100 using ResNet-32.}
  \centering 
  \resizebox{5.25in}{!}{%
\hspace{-0.15in}  \begin{tabular}{cccc}
    \toprule     
    \textbf{Algorithm}&\textbf{\textcolor{black}{Squared bias}}&\textbf{Variance}&\textbf{\textcolor{black}{Accuracy (\%)}}\\
    \midrule
    Vanilla &$0.69$&$0.49$&$40.8$ \\
    LDAM+DRW \cite{NEURIPS2019_621461af} &$0.55$&$0.57$&$43.8$\\
    cRT \cite{kang2019decoupling}&$0.59$&$0.52$&$44.3$\\
    RIDE \cite{wang2021longtailed}&$0.59$&$0.43$&$48.9$\\
    \midrule
    DAMEL without using EMA weights&$0.56$&$0.45$&$49.5$ \\
    DAMEL without aggregating representations&$0.54$&$0.46$&$50.0$ \\
    \textbf{DAMEL with 3 experts}&$\mathbf{0.53}$&$\mathbf{0.42}$&$\mathbf{52.6}$ \\
    \bottomrule
  \end{tabular}}
\label{biasvariancetable}
\end{table} 

\textcolor{black}{\subsection{Comparison with Iteration-Level EMA}}
\label{sec:iteration_epoch_ema}

\textcolor{black}{To further analyze the proposed time-axis ensemble, we compare our epoch-level EMA with \textcolor{black}{the} conventional iteration-level EMA. As shown in Table~\ref{tab:iteration_epoch_ema}, epoch-level EMA consistently outperforms iteration-level EMA on both CIFAR-10-LT and CIFAR-100-LT \textcolor{black}{across} different \textcolor{black}{class-}imbalance ratios. \textcolor{black}{These results indicate} that the \textcolor{black}{EMA} update frequency \textcolor{black}{plays an} important \textcolor{black}{role in} constructing an effective time-axis ensemble.}

\begin{table}[htbp]
  \caption{\textcolor{black}{Prediction accuracy (\%) using} iteration-level EMA and epoch-level EMA.}
  \centering
  \begin{tabular}{c|cc|cc}
    \toprule
    \multirow{2}{*}{\textbf{EMA update frequency}} 
    & \multicolumn{2}{c|}{\textbf{CIFAR-10-LT}} 
    & \multicolumn{2}{c}{\textbf{CIFAR-100-LT}} \\
    \cline{2-5}
    & $\gamma=100$ & $\gamma=50$ 
    & $\gamma=100$ & $\gamma=50$ \\
    \midrule
    Iteration-level EMA 
    & 81.7 & 84.4 & 48.5 & 53.2 \\
    Epoch-level EMA (Ours) 
    & \textbf{85.0} & \textbf{87.9} & \textbf{53.1} & \textbf{57.8} \\
    \bottomrule
  \end{tabular}
  \label{tab:iteration_epoch_ema}
\end{table}
\textcolor{black}{Iteration-level EMA updates the averaged weights after every mini-batch, \textcolor{black}{when} consecutive model \textcolor{black}{states} are highly correlated. \textcolor{black}{Consequently}, aggregating these highly similar weights may provide \textcolor{black}{only} limited ensemble diversity. In contrast, epoch-level EMA updates the averaged weights only after the model has processed the entire training set, \textcolor{black}{resulting in} temporally more separated and diverse model states.}

\textcolor{black}{This \textcolor{black}{approach} is particularly \textcolor{black}{beneficial for CIL}, where mini-batch-level class compositions can \textcolor{black}{fluctuate substantially} under severe imbalance. Updating \textcolor{black}{the} EMA at the epoch level reduces the influence of such mini-batch\textcolor{black}{-level} fluctuations and \textcolor{black}{better} captures \textcolor{black}{the overall} training trajectory. As a result, DAMEL \textcolor{black}{constructs} a more effective time-axis ensemble that reduces prediction variance while preserving \textcolor{black}{the representations learned during} later \textcolor{black}{stages of training}.}

\textcolor{black}{\subsection{Computational Cost Analysis}}
\label{sec:computational_cost}
\textcolor{black}{We analyze the computational cost of DAMEL by varying the number of experts\textcolor{black}{,} as shown in Table~\ref{tab:computational_cost2}. We report \textcolor{black}{floating} point \textcolor{black}{operations} (FLOPs), GPU memory usage, the number of trainable parameters, and inference time \textcolor{black}{for} CIFAR-10-LT, CIFAR-100-LT, and ImageNet-LT. All measurements were \textcolor{black}{obtained using} an NVIDIA GeForce RTX 3090 Ti GPU.}

\begin{table}[htbp]
  \centering
  \caption{Computational cost analysis of DAMEL with different numbers of experts\textcolor{black}{.}}
  \centering\resizebox{5.5in}{!}{
  \begin{tabular}{c|c|c|c|c|c}
    \toprule
    \textbf{Dataset} & \textbf{\# Experts} & \textbf{FLOPs} & \textbf{GPU memory} & \textbf{\# Params} & \textbf{Inference time} \\
    & & \textbf{(iter/s)} & \textbf{(MiB)} & & \textbf{(s)} \\
    \midrule
    \multirow{4}{*}{CIFAR-10-LT}
    & 1 & 46.5 & 1728 & 464,804 & 1.17 \\
    & 2 & 40.4 & 1746 & 817,582 & 1.35 \\
    & 3 & 35.5 & 1760 & 1,170,360 & 1.42 \\
    & 4 & 32.0 & 1778 & 1,523,138 & 1.47 \\
    \midrule
    \multirow{4}{*}{CIFAR-100-LT}
    & 1 & 44.7 & 1728 & 467,504 & 1.28 \\
    & 2 & 38.2 & 1748 & 840,892 & 1.35 \\
    & 3 & 34.2 & 1760 & 1,205,280 & 1.43 \\
    & 4 & 30.1 & 1778 & 1,569,668 & 1.45 \\
    \midrule
    \multirow{2}{*}{ImageNet-LT}
    & 3 & 2.67 & 15018 & 65,729,504 & 51.56 \\
    & 4 & 2.54 & 15658 & 84,791,240 & 50.73 \\
    \bottomrule
  \end{tabular}}
  \label{tab:computational_cost2}
\end{table}

\textcolor{black}{As the number of experts increases, the number of trainable parameters increases, \textcolor{black}{whereas} the FLOPs gradually decrease across all datasets. \textcolor{black}{These results illustrate the} practical performance\textcolor{black}{-}-cost trade-off \textcolor{black}{of DAMEL}: increasing the number of experts can \textcolor{black}{enhance} representation diversity and predictive performance while \textcolor{black}{incurring} additional computational \textcolor{black}{overhead}. Therefore, the number of experts can be selected according to the desired balance between \textcolor{black}{predictive} accuracy and \textcolor{black}{computational} efficiency.}



\textcolor{black}{\subsection{Capacity-controlled ablation study}}
\textcolor{black}{To \textcolor{black}{determine} whether the \textcolor{black}{performance gains} of DAMEL \textcolor{black}{arise} from expert diversity rather than \textcolor{black}{merely} from increased representation dimensionality, we conducted a capacity-controlled ablation study. \textcolor{black}{Specifically, we} constructed a single-expert model whose representation dimension matches \textcolor{black}{that of} the concatenated representation used in DAMEL. As shown in Table~\ref{tab:capacity_controlled_ablation}, this model consistently underperforms DAMEL on \textcolor{black}{both} CIFAR-10-LT and CIFAR-100-LT, indicating that the improvement \textcolor{black}{cannot be attributed solely} to increased feature dimensionality or model capacity. Instead, the \textcolor{black}{performance} gain \textcolor{black}{stems} from combining diverse expert-specific representations. As \textcolor{black}{illustrated} in Figure~\ref{differentrep} and Appendix Figure~5, different experts attend to \textcolor{black}{distinct} discriminative regions and learn complementary features\textcolor{black}{. By concatenating these representations}, the auxiliary balanced classifier \textcolor{black}{can} jointly exploit this complementary information.}
\begin{table}[htbp]
  \caption{\textcolor{black}{Prediction accuracy (\%) in the} capacity-controlled ablation study. The capacity-controlled model increases only the representation dimension without using multiple expert branches.}
  \centering\resizebox{5in}{!}{
  \begin{tabular}{c|cc|cc}
    \toprule
    \multirow{2}{*}{\textbf{Method}} 
    & \multicolumn{2}{c|}{\textbf{CIFAR-10-LT}} 
    & \multicolumn{2}{c}{\textbf{CIFAR-100-LT}} \\
    \cline{2-5}
    & $\gamma=100$ & $\gamma=50$ 
    & $\gamma=100$ & $\gamma=50$ \\
    \midrule
    Enlarged representation only 
    & 83.6 & 86.4 & 51.3 & 56.4 \\
    DAMEL (4 experts)
    & \textbf{85.0} & \textbf{87.9} & \textbf{53.1} & \textbf{57.8} \\
    \bottomrule
  \end{tabular}}
  \label{tab:capacity_controlled_ablation}
\end{table}

\textcolor{black}{\subsection{Sensitivity Analysis}}
\label{sec:sensitivity_analysis}

\textcolor{black}{To provide a more systematic \textcolor{black}{evaluation} of DAMEL, we \textcolor{black}{conducted} sensitivity \textcolor{black}{analyses} on four key hyperparameters: the number of experts, the EMA update hyperparameter $\beta$, the classifier scale, and the class-balanced loss coefficient $\lambda_{\mathrm{CB}}$. The results are summarized in Tables~\ref{tab:sensitivity_experts}--\ref{tab:sensitivity_cb}.}

\textcolor{black}{As shown in Table~\ref{tab:sensitivity_experts}, increasing the number of experts generally improves the performance of DAMEL. This \textcolor{black}{finding} supports our \textcolor{black}{hypothesis} that multiple experts \textcolor{black}{learn} more diverse representations, which can be effectively \textcolor{black}{leveraged} by the auxiliary classifier.} 

\begin{table}[htbp]
  \caption{\textcolor{black}{Prediction accuracy (\%) for different numbers of experts.}}
  \centering
  \begin{tabular}{c|cc|cc|c}
    \toprule
    \multirow{2}{*}{\textbf{\# Experts}}
    & \multicolumn{2}{c|}{\textbf{CIFAR-10-LT}}
    & \multicolumn{2}{c|}{\textbf{CIFAR-100-LT}}
    & \textbf{ImageNet-LT} \\
    \cline{2-6}
    & $\gamma=100$ & $\gamma=50$
    & $\gamma=100$ & $\gamma=50$
    & ResNet-50 \\
    \midrule
    1 & 81.8 & 84.9 & 50.0 & 53.8 & - \\
    2 & 82.1 & 85.5 & 52.1 & 56.3 & - \\
    3 & 84.7 & 86.5 & 52.6 & 56.9 & 56.8 \\
    4 & \textbf{85.0} & \textbf{87.9} & \textbf{53.1} & \textbf{57.8} & \textbf{57.7} \\
    \bottomrule
  \end{tabular}
  \label{tab:sensitivity_experts}
\end{table}

\textcolor{black}{As shown in Table~\ref{tab:sensitivity_ema}, the EMA update hyperparameter $\beta$ has a \textcolor{black}{noticeable} effect on performance. When $\beta$ is too small, such as $\beta=0.01$, the EMA weights are updated too slowly and \textcolor{black}{may} not \textcolor{black}{adequately} reflect recently learned network \textcolor{black}{parameters}. In contrast, when $\beta$ is too large, such as $\beta=0.3$, the EMA \textcolor{black}{places excessive emphasis on} recent weights\textcolor{black}{, reducing} the stabilizing \textcolor{black}{effect} of temporal weight aggregation. The results \textcolor{black}{indicate} that moderate values, such as $\beta=0.05$ or $\beta=0.1$, provide a \textcolor{black}{favorable} balance.}

\begin{table}[htbp]
  \caption{\textcolor{black}{Prediction accuracy (\%) for different values of the EMA update hyperparameter $\beta$.}}
  \centering
  \begin{tabular}{c|cc|cc}
    \toprule
    \multirow{2}{*}{\textbf{EMA $\beta$}}
    & \multicolumn{2}{c|}{\textbf{CIFAR-10-LT}}
    & \multicolumn{2}{c}{\textbf{CIFAR-100-LT}} \\
    \cline{2-5}
    & $\gamma=100$ & $\gamma=50$
    & $\gamma=100$ & $\gamma=50$ \\
    \midrule
    0.01 & 81.1 & 83.3 & 49.0 & 54.5 \\
    0.05 & \textbf{84.8} & \textbf{86.9} & 52.4 & \textbf{57.0} \\
    0.10 & 84.7 & 86.5 & \textbf{52.6} & 56.9 \\
    0.20 & 84.1 & 86.4 & 51.4 & 56.0 \\
    0.30 & 83.3 & 85.6 & 50.1 & 55.0 \\
    \bottomrule
  \end{tabular}
  \label{tab:sensitivity_ema}
\end{table}

\textcolor{black}{Table~\ref{tab:sensitivity_scale} shows that DAMEL is \textcolor{black}{relatively insensitive} to the classifier scale within a reasonable range. In particular, scale values between 16 and 24 \textcolor{black}{yield} stable performance across datasets and \textcolor{black}{class-}imbalance ratios. This \textcolor{black}{result} indicates that DAMEL does not \textcolor{black}{depend} on a \textcolor{black}{narrowly tuned} classifier scale.}

\begin{table}[htbp]
  \caption{\textcolor{black}{Prediction accuracy (\%) for different classifier scales.}}
  \centering
  \begin{tabular}{c|cc|cc}
    \toprule
    \multirow{2}{*}{\textbf{Scale}}
    & \multicolumn{2}{c|}{\textbf{CIFAR-10-LT}}
    & \multicolumn{2}{c}{\textbf{CIFAR-100-LT}} \\
    \cline{2-5}
    & $\gamma=100$ & $\gamma=50$
    & $\gamma=100$ & $\gamma=50$ \\
    \midrule
    8  & 84.4 & \textbf{86.8} & 50.7 & 54.2 \\
    16 & 84.4 & 86.5 & \textbf{52.6} & 56.4 \\
    20 & \textbf{84.7} & 86.5 & \textbf{52.6} & \textbf{56.9} \\
    24 & 84.2 & 86.2 & 52.1 & 56.7 \\
    \bottomrule
  \end{tabular}
  \label{tab:sensitivity_scale}
\end{table}

\textcolor{black}{Finally, Table~\ref{tab:sensitivity_cb} shows that the class-balanced loss coefficient $\lambda_{\mathrm{CB}}$ also \textcolor{black}{has a limited effect on performance} within the tested range. DAMEL maintains stable performance across moderate variations \textcolor{black}{in} $\lambda_{\mathrm{CB}}$, \textcolor{black}{demonstrating robustness} to the strength of the auxiliary class-balanced loss.} 

\begin{table}[htbp]
  \caption{\textcolor{black}{Prediction accuracy (\%) for different values of the class-balanced loss coefficient $\lambda_{\mathrm{CB}}$.}}
  \centering
  \begin{tabular}{c|cc|cc}
    \toprule
    \multirow{2}{*}{$\boldsymbol{\lambda_{\mathrm{CB}}}$}
    & \multicolumn{2}{c|}{\textbf{CIFAR-10-LT}}
    & \multicolumn{2}{c}{\textbf{CIFAR-100-LT}} \\
    \cline{2-5}
    & $\gamma=100$ & $\gamma=50$
    & $\gamma=100$ & $\gamma=50$ \\
    \midrule
    0.5 & 84.3 & 86.9 & 52.3 & 56.7 \\
    1.0 & \textbf{84.7} & 86.5 & \textbf{52.6} & \textbf{56.9} \\
    1.5 & 84.4 & \textbf{87.2} & 52.5 & 56.7 \\
    2.0 & 84.3 & 86.8 & 52.1 & 56.7 \\
    \bottomrule
  \end{tabular}
  \label{tab:sensitivity_cb}
\end{table}

\textcolor{black}{Overall, these analyses \textcolor{black}{demonstrate} that DAMEL performs consistently across reasonable hyperparameter ranges \textcolor{black}{and provide} practical \textcolor{black}{guidance} for selecting the number of experts, EMA \textcolor{black}{update hyperparameter} $\beta$, classifier scale, and \textcolor{black}{class-balanced loss coefficient} $\lambda_{\mathrm{CB}}$.}

\subsection{Ablation Study}
\label{ablation}

To investigate the \textcolor{black}{contribution} of each \textcolor{black}{component} of DAMEL, we conducted a \textcolor{black}{fine-grained}  ablation study on CIFAR-100-LT with $\gamma=100$. The results are presented in Table \ref{ablationtable} \textcolor{black}{and can be} summarized as follows. \textbf{\textcolor{black}{(}1)} DAMEL with all \textcolor{black}{components} achieved the highest accuracy. \textbf{\textcolor{black}{(}2)} \textcolor{black}{Removing} the EMA weights \textcolor{black}{reduced} prediction accuracy. \textbf{\textcolor{black}{(}3)} \textcolor{black}{Removing representation aggregation also} decreased prediction accuracy. \textbf{\textcolor{black}{(4)}} 
\textcolor{black}{Training the auxiliary classifier without a class-balanced loss led to a substantial drop in performance.} \textbf{\textcolor{black}{(5)}} \textcolor{black}{Averaging the representation of multiple experts, rather than concatenating them, reduced prediction accuracy.} \textbf{\textcolor{black}{(}6)} \textcolor{black}{Updating} the EMA weights \textcolor{black}{at every} iteration\textcolor{black}{,} rather than \textcolor{black}{at the end of} each epoch\textcolor{black}{, resulted in lower performance}. \textbf{\textcolor{black}{(}7)} \textcolor{black}{Replacing EMA with} simple \textcolor{black}{weight} averaging\textcolor{black}{, i.e.,} SWA, \textcolor{black}{also decreased} prediction accuracy. \textbf{\textcolor{black}{(}8)} \textcolor{black}{Using either} a \textcolor{black}{larger} or \textcolor{black}{smaller} EMA update hyperparameter\textcolor{black}{, such as} $\beta=0.3$ \textcolor{black}{or} $\beta=0.01$, \textcolor{black}{reduced performance. Across} repeated experiments with \textcolor{black}{different} $\beta$ values, we found that $\beta$ values between 0.05 and 0.2 \textcolor{black}{provided effective} EMA \textcolor{black}{updates}. \textbf{\textcolor{black}{(}9)} \textcolor{black}{Decoupling the training} of the auxiliary classifier \textcolor{black}{from the learning of the expert} representations \textcolor{black}{also decreased} prediction accuracy. \textcolor{black}{Collectively,} these results \textcolor{black}{confirm} that each \textcolor{black}{component} of DAMEL \textcolor{black}{contributes to its overall effectiveness}. 

\begin{table}[htbp]
  \caption{\textcolor{black}{Prediction accuracy (\%) of DAMEL under different ablation settings on CIFAR-100-LT with $\gamma=100$.}}
  \centering\resizebox{5.45in}{!}{
  \begin{tabular}{l|c|c|c|c}
    \toprule
    \textbf{Ablation study} & \textbf{Overall} & \textcolor{black}{\textbf{Many}} & \textcolor{black}{\textbf{Medium}} & \textcolor{black}{\textbf{Few}} \\
    \midrule
    DAMEL with 3 experts (proposed algorithm) & 52.6 & 70.4 & 53.5 & 31.5 \\
    \midrule
    Without using EMA weights & 49.5 & 70.6 & 50.0 & 25.0 \\
    Without aggregating representations (1 expert) & 49.2 & 64.6 & 50.6 & 29.0 \\
    Without using class-balanced loss & 46.4 & 75.3 & 46.4 & 13.4 \\
    With averaging the representations & 50.0 & 66.8 & 50.1 &  28.2\\
    With updating EMA weights for each iteration & 48.5 & 62.4 & 50.3 & 28.8 \\
    With simple averaging of weights (SWA) & 50.7 & 66.8 & 52.6 & 30.3 \\
    With higher EMA update hyperparameter ($\beta=0.3$) & 50.2 & 68.5 & 50.2 & 25.9 \\
    With lower EMA update hyperparameter ($\beta=0.01$) & 49.6 & 68.3 & 49.4 & 28.7 \\
    With decoupled learning of auxiliary classifier & 46.7 & 67.4 & 44.5 & 25.5 \\
    \bottomrule
  \end{tabular}}
  \label{ablationtable}
\end{table}

To further \textcolor{black}{examine} the \textcolor{black}{contribution} of EMA \textcolor{black}{weight aggregation}, we also conducted experiments on CIFAR-10-LT and CIFAR-100-LT with and without EMA weights while \textcolor{black}{varying} the number of experts\textcolor{black}{. The corresponding results are provided} in Appendix G.

\section{Conclusion}

We proposed DAMEL, a new CIL algorithm \textcolor{black}{that employs} multiple experts along \textcolor{black}{two complementary} axes to \textcolor{black}{reduce} both \textcolor{black}{prediction} bias and variance. \textcolor{black}{Along the representation axis,} DAMEL \textcolor{black}{aggregates representations learned by multiple experts, enabling} the auxiliary classifier to \textcolor{black}{leverage diverse and complementary features}. \textcolor{black}{Along} the time axis\textcolor{black}{, DAMEL aggregates network weights} by updating the EMA weights at the end of each epoch. Experimental results across various \textcolor{black}{datasets and class-imbalance settings} demonstrate \textcolor{black}{that DAMEL consistently outperforms existing} multi-expert learning and CIL algorithms. \textcolor{black}{Extensive} qualitative analysis and ablation studies \textcolor{black}{further confirm the effectiveness and contribution of} each component of DAMEL. \textcolor{black}{One limitation of} this study \textcolor{black}{is that} hyperparameters \textcolor{black}{were selected} heuristically based on experimental results \textcolor{black}{rather than} theoretical \textcolor{black}{principles}. In future work, we aim to \textcolor{black}{develop a} theoretical \textcolor{black}{framework} for determining the optimal number of experts and the optimal EMA update hyperparameter. 

\section*{Conflict of Interest Disclosure}
The authors declare that they have no conflict of interests that could have appeared to influence the work reported in this manuscript.

\section*{Acknowledgments}
The authors would like to thank the reviewers and the
editor for reviewing this article and providing valuable comments. This work was supported by the National Research Foundation of Korea (NRF) grant funded by the Korea government (MSIT) (2023R1A2C2005453, RS-2023-00218913).

\clearpage
\newpage




\bibliographystyle{elsarticle-harv} 
\bibliography{reference}

\end{document}